\documentclass[conference]{IEEEtran}
\IEEEoverridecommandlockouts
\usepackage{cite}
\usepackage{amsmath,amssymb,amsfonts}
\usepackage{graphicx}
\usepackage{textcomp}
\usepackage{xcolor}
\usepackage{booktabs}
\usepackage{multirow}
\usepackage{siunitx}
\usepackage{algorithm}
\usepackage{algpseudocode}
\usepackage{threeparttable}
\usepackage[hidelinks]{hyperref} 

\def\BibTeX{{\rm B\kern-.05em{\sc i\kern-.025em b}\kern-.08em
    T\kern-.1667em\lower.7ex\hbox{E}\kern-.125emX}}
\begin{document}

\title{Location-Based Safety Management with Wearable Seamless Indoor–Outdoor Positioning

\thanks{This research was funded by the National Fire Agency (NFA) and the Korea Planning \& Evaluation Institute of Industrial Technology (KEIT) of the Republic of Korea (No. RS-2025-02313957).}
}




\author{
\IEEEauthorblockN{Anh-Van Vu, Seungwoo Chae, Changmin Sung, Van-Tuan Nguyen and Dongsoo Han}
\IEEEauthorblockA{
\textit{Department of Computer Science, KAIST} \\
Daejeon, Republic of Korea \\
\{vuanhvan, shatie73, shade, tuan.nv, dshan\}@kaist.ac.kr
}
}
\maketitle

\begin{abstract}

Reliable worker localization is essential for real-time industrial safety management, but seamless outdoor–indoor positioning remains challenging. GNSS performs well outdoors yet degrades near buildings and fails indoors, while many indoor systems require costly infrastructure or extensive calibration. This paper presents a practical safety-oriented positioning system that uses resource-constrained wearable tags and a device--cloud split architecture for continuous outdoor--indoor worker tracking. The tag integrates GNSS, IMU, barometric sensing, LTE Cat-M communication, and performs lightweight sensor fusion, while the cloud server executes map-aware fusion for continuous tracking. Developed as an operational platform, the system also includes a database, a map tool, and a dashboard that provide real-time monitoring, geofencing-based access alerts, and safety and productivity analysis. Experiments with multiple participants, each walking an approximately 1400 m outdoor–indoor route, using LTE Cat-M communication, achieve mean errors of 1.32 m indoors and 3.30 m outdoors, while maintaining low positioning latency and efficient power consumption. These results demonstrate the feasibility of wearable-based seamless positioning for scalable real-time safety management.

\end{abstract}


\section{Introduction}

In industrial worksites and large-scale facilities, worker localization is a fundamental requirement for safety management. Real-time location information allows managers to monitor worker distribution, detect entry into hazardous or restricted zones, support emergency response, and analyze mobility patterns for safer and more efficient operations \cite{chung2023iot, Nguyen2026EdgeCentric, jin2020iot}. However, seamless pedestrian positioning across outdoor and indoor spaces remains challenging. GNSS offers broad outdoor coverage but degrades near buildings and often becomes unavailable indoors. Meanwhile, many indoor positioning systems depend on BLE\cite{chen2021data, hu2022experience}, UWB\cite{mayer2023self, zhang2025seamless}, or RFID\cite{chung2023iot} infrastructure, which can be costly to deploy and maintain, especially in dynamic workplaces, temporary sites, and large facilities with frequently changing layouts.

A practical safety-management system therefore needs an infrastructure-light, scalable, and continuously available positioning solution that works across outdoor and indoor spaces. Wearable sensing is a promising direction because each worker can wear a compact device that collects inertial, barometric, and GNSS measurements without dense localization infrastructure. However, wearable tags are constrained by size, power, computation, memory, and communication bandwidth. Running complex map-aware fusion entirely on the device can exceed these limits, while streaming raw sensor data to a remote server increases energy consumption, latency, and network load. Practical wearable positioning thus requires a careful split between local device and server-side processing.

This paper presents a wearable seamless pedestrian positioning system for location-based safety management (LBSM). The system applies a distributed architecture in which the wearable tag performs sensing and lightweight local processing to produce motion summaries, while a cloud server fuses these summaries with GNSS, barometric pressure, and indoor occupancy maps using an event-driven map-aware particle filter. The system further integrates map management, real-time monitoring, geofencing, and safety analytics to support practical LBSM deployment. This system demonstrates how resource-aware wearable sensing and server-side intelligence can jointly support scalable, real-time positioning for practical safety management.


The main contributions of this work are as follows:
\begin{itemize}
    \item We develop an end-to-end, infrastructure-light seamless pedestrian positioning system using resource-constrained wearable tags and a device--cloud split architecture, enabling reliable and scalable LBSM. The system integrates custom LTE Cat-M wearable tags, a positioning server, operational and map databases, a map tool, and a real-time dashboard.

    \item We develop the operational LBSM functions built upon the positioning system, including real-time worker and device monitoring, geofencing-based access alerts, and location-based safety and productivity analytics.

    \item We validate the complete system with multiple participants, each traversing an approximately 1400~m outdoor--indoor route, and evaluate its positioning accuracy, communication latency under different IoT protocols and connectivity conditions, wearable power consumption, and geofencing-based alert performance.
\end{itemize}

\section{Related Work}

Pedestrian localization increasingly relies on heterogeneous sensing, combining inertial motion with radio, map, and environmental cues \cite{chen2021data,hu2022experience,zhang2025seamless,wang2023seamless,vu2022iot}. For example, Chen et al. improve accuracy by learning user motion from smartphone IMU data and correcting drift using BLE measurements and particle filtering~\cite{chen2021data}, while MLoc demonstrates large-scale positioning services by integrating BLE, IMU, and geomagnetic sensing within an edge-assisted architecture~\cite{hu2022experience}. To improve accuracy while reducing operating costs, Mayer et al. develop self-sustaining UWB tags powered by solar cells~\cite{mayer2023self}. For seamless localization, Zhang et al. fuse GNSS, IMU, UWB, and map information using FGO ~\cite{zhang2025seamless}, whereas Wang et al. combine foot-mounted inertial sensing with smartphone PPP, a 3-D map, and a barometer using an EKF~\cite{wang2023seamless}. Despite these advances, existing approaches often depend on dense infrastructure, access to smartphone sensors, substantial local computation, or controlled deployment conditions. Providing scalable, reliable, and continuous outdoor--indoor positioning with resource-constrained wearables thus remains challenging.

LBSM has developed in parallel. IoT-based monitoring systems use wearable or mobile devices, cloud platforms, and access-control logic to detect unsafe situations and improve manager awareness~\cite{chung2023iot,jin2020iot}. Recent RTLS- and vision-based studies provide proximity alerts around equipment and support real-time hazard recognition~\cite{huang2021providing, jeelani2021real}. However, these systems often treat localization as a supporting component and depend on expensive RFID/BLE/UWB infrastructure or cameras deployed around the worksite. A gap therefore remains between infrastructure-light seamless positioning and its integration into a complete operational safety-management platform. This work addresses this gap by integrating wearable positioning into an end-to-end LBSM platform. The tag and server jointly provide seamless outdoor--indoor positioning using wearable sensing and map constraints, while the surrounding platform supports real-time worker and device monitoring, geofencing-based access alerts, and location-based safety and productivity analytics. This integration enables practical safety management under limited infrastructure and connectivity across heterogeneous environments.

\section{System Design}

\subsection{System Architecture}

\begin{figure}[t]
  \centering
  \includegraphics[width= 85 mm]{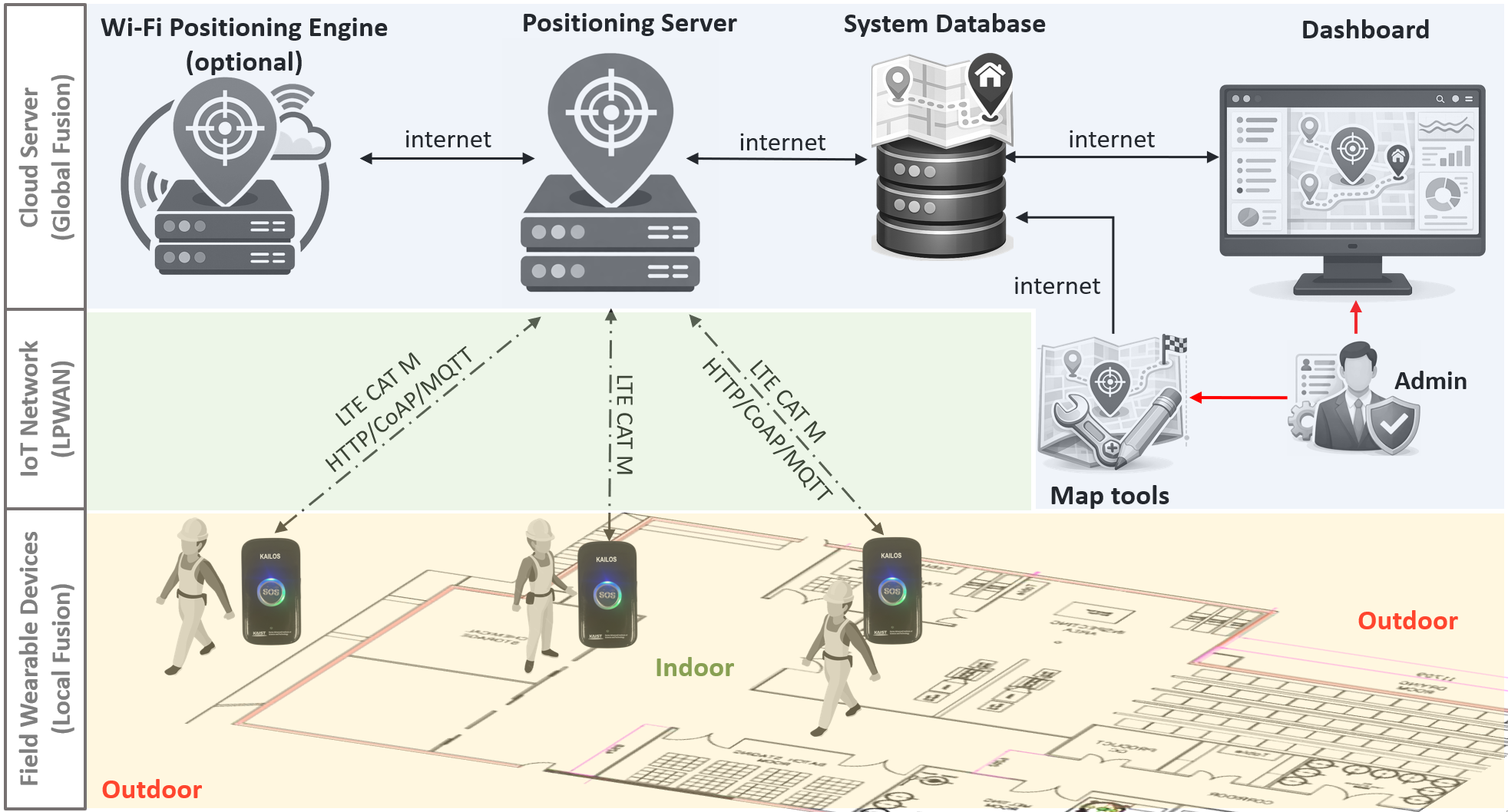}
  \caption{System architecture of the proposed wearable positioning platform.}
  \label{fig:system_architecture}
\end{figure}

The system is designed as an end-to-end positioning platform for LBSM using resource-constrained wearables. To meet the stringent requirements of positioning accuracy, latency, and efficiency, its architecture integrates five main components: wearable tags, a positioning server, a database, a dashboard server, and a map tool. An external Wi-Fi positioning engine \cite{han2014kailos, vu2022iot} may also be incorporated as an optional positioning cue, although it is excluded from the evaluation in this study. The architecture distributes processing to balance computational demand with hardware efficiency, while providing accurate and responsive position estimates. Each tag performs local sensing, lightweight fusion, and communication, while the positioning server handles computationally intensive sensor fusion using device data and environmental information stored in the map database. The tags communicate with the server over low-bandwidth links using selectable IoT protocols, including CoAP, HTTP, and MQTT. This division enables the use of a sophisticated Map-Aware Particle Filter without exceeding the wearable device’s resource constraints. The map tool supports environment preparation and management, while the dashboard provides real-time monitoring and service operation. The overall architecture is illustrated in Fig.~\ref{fig:system_architecture}.

\subsection{Wearable Tag}
\subsubsection{Hardware Design}

\begin{figure}[b!]
  \centering
  \includegraphics[width= 75 mm]{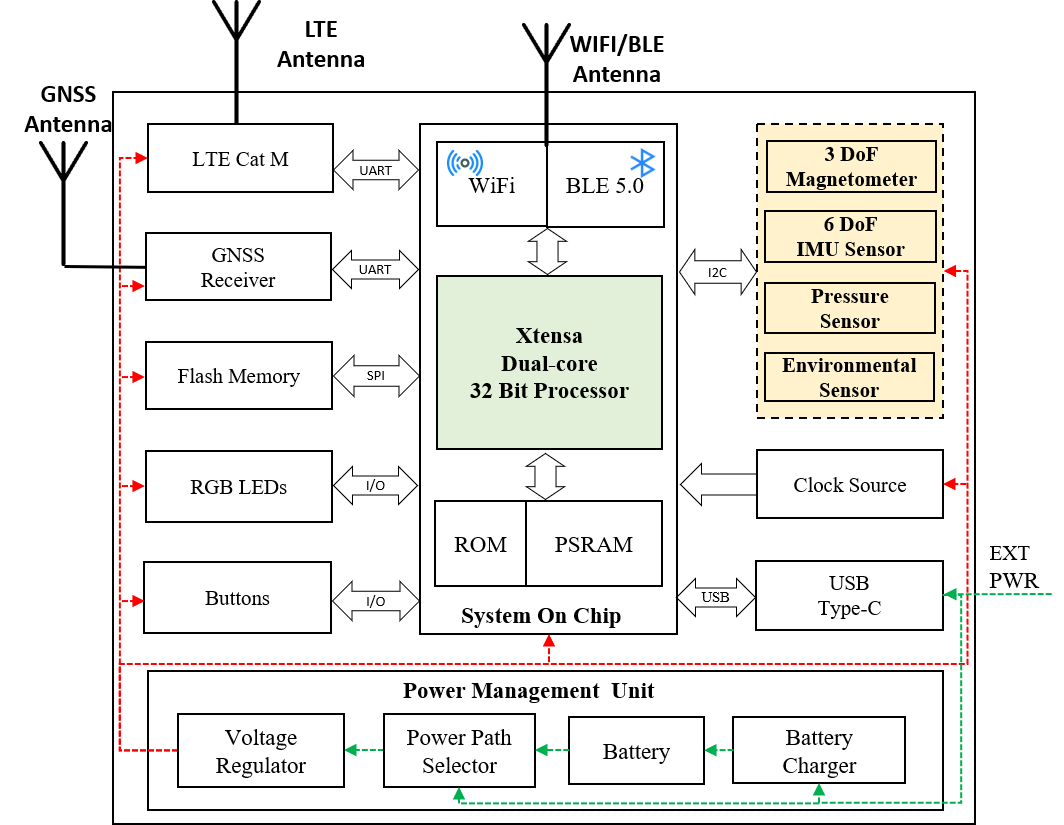}
  \caption{Hardware block diagram of the wearable IoT device.}
  \label{fig: hardware_block_diagram}
\end{figure}

\begin{figure}[t!]
  \centering
  \includegraphics[width=79mm]{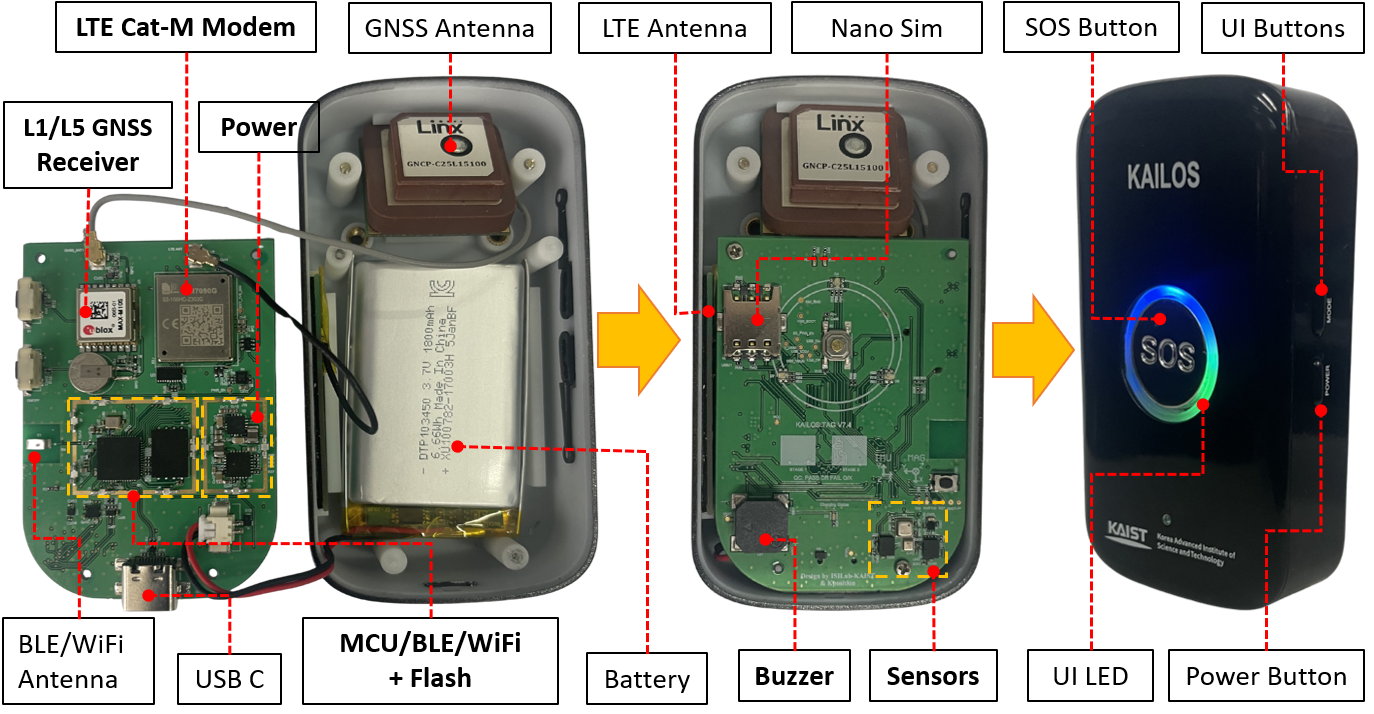}
  \caption{Components of the custom wearable IoT device.}
  \label{fig:hard_ware_breakdown}
\end{figure}

The wearable platform is designed for LBSM under strict size, power, and cost constraints, while supporting continuous sensing during daily worker activities. As shown in Fig.~\ref{fig: hardware_block_diagram}, the system is built around a 32-bit dual-core processor that coordinates sensor acquisition, on-board processing, wireless communication, memory, interfaces, and power control. The processor integrates 384~KB ROM, 2~MB PSRAM, Wi-Fi, and BLE 5.0 on a single chip, while a 128~MB SPI NOR flash supports firmware and application data storage, and an LTE Cat-M modem connected via UART provides low-power IoT communication with a maximum data rate of 1119~kbps uplink and 589~kbps downlink.

For positioning and context awareness, the device integrates a dual-band L1/L5 GNSS receiver and an I\textsuperscript{2}C-based sensor suite. The 6-DoF IMU combines a 16-bit triaxial accelerometer and gyroscope for motion estimation, while a 3-DoF magnetometer provides geomagnetic orientation cues (not used in this study). A barometric pressure sensor supports altitude estimation with $\pm 6~\mathrm{Pa}$ relative accuracy, and an environmental sensor captures temperature, humidity, air quality, and volatile organic compound levels for richer monitoring capability.

Power management is handled through a protected switching converter for the main 3.3~V rail and an ultra-low-dropout regulator for the 1.8~V sensor rail, reducing thermal loss and improving sensor stability. An integrated charger enables fast charging of the 1800~mAh battery. A USB Type-C port provides charging and data access, while buttons and an LED enable basic user interaction and status indication. The device breakdown is shown in Fig.~\ref{fig:hard_ware_breakdown}.


\subsubsection{Software Design}

\begin{figure}[b!]
  \centering
  \includegraphics[width= 80 mm]{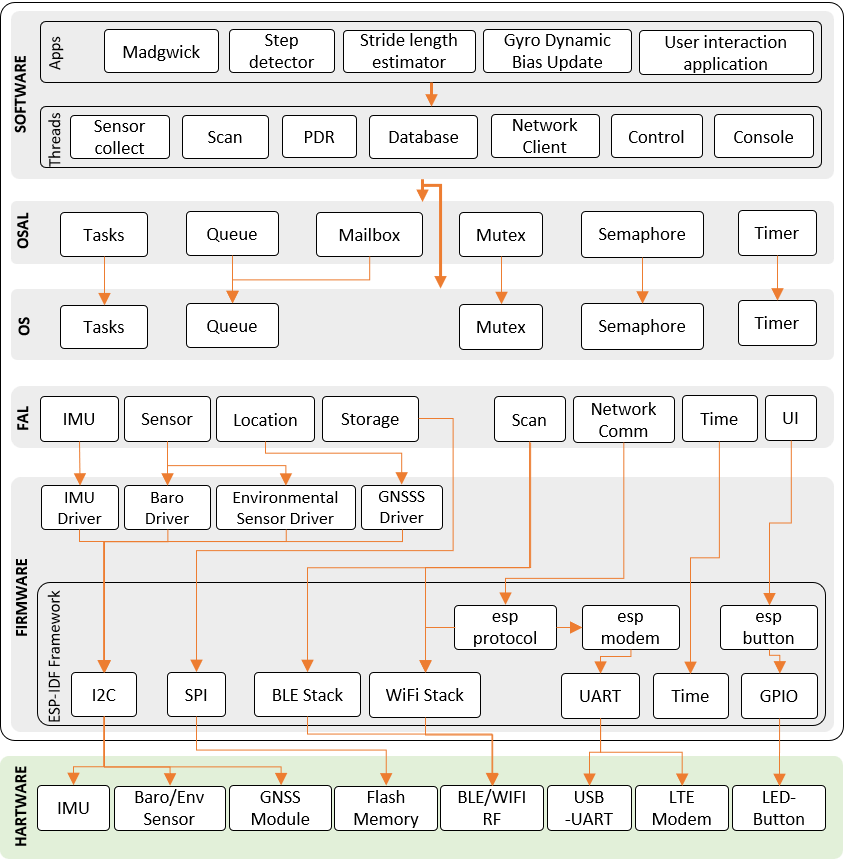}
  \caption{Software architecture of the IoT device.}
  \label{fig:software-structure}
\end{figure}

The device software follows a layered, object-oriented architecture to support modularity, portability, and real-time operation. As shown in Fig.~\ref{fig:software-structure}, it consists of five layers: the \textit{firmware}, \textit{firmware abstraction layer (FAL)}, \textit{operating system (OS)}, \textit{operating system abstraction layer (OSAL)}, and \textit{software layer}. This separation isolates hardware-specific functions from application logic, reducing redesign effort when hardware is changed. At the lowest level, the \textit{firmware layer} provides drivers for sensors, flash memory, communication interfaces, and network protocols using the Espressif IoT Development Framework. It enables low-level access to radio modems, sensors, and memory. The \textit{firmware abstraction layer} wraps these drivers into unified APIs for sensing, communication, logging, and timing, allowing upper layers to access hardware resources through consistent interfaces. The \textit{operating system layer} is based on FreeRTOS with symmetric multiprocessing (SMP), enabling concurrent execution on the dual-core processor. Tasks, queues, timers, and semaphores support real-time scheduling and synchronization. The \textit{OS abstraction layer} further standardizes these primitives and provides lightweight mailboxes for efficient inter-task communication with reduced memory overhead. The top \textit{software layer} implements the main application logic through reusable libraries and event-driven tasks. General libraries include deterministic memory pools and a CRC-protected serial packet protocol for reliable, low-latency communication. Device functions are organized into tasks for sensor acquisition, on-board PDR processing, radio scanning, storage, server communication, system control, and console debugging. Each task runs as an event-driven state machine with its own configuration, priority, and communication channels, remaining idle until triggered to improve real-time predictability.


\subsubsection{On-board PDR}
On-board PDR is a core function of the wearable IoT device, especially under bandwidth-constrained communication. It transforms high-rate IMU data into compact, low-rate motion increments, which greatly reduce communication load while retaining the information needed for cloud-based map-aware fusion. The detailed design and implementation of the on-board PDR are presented in our previous work~\cite{vu2024real, vu2025black}. Algorithm~\ref{alg:onboard_pdr} illustrates the main pipeline of the on-board PDR implemented on the wearable tag. Here, $\mathcal S(\cdot)$ detects stance onsets, $\mathcal G(\cdot)$ estimates gyroscope bias, $\mathcal H(\cdot)$ updates orientation with bias-corrected AHRS, $\Psi(\cdot)$ extracts heading, and $\mathcal D(\cdot)$ estimates stride length. The resulting motion summary is $\mathbf u_m=(s_m,\psi_{\tau_m})$, with $\mathcal U=\{\mathbf u_m\}$.

\begin{algorithm}[t!]
\caption{On-board PDR Motion Summary}
\label{alg:onboard_pdr}
\resizebox{75mm}{!}{%
\begin{minipage}{\linewidth}
\begin{algorithmic}[1]
\Require IMU stream $\{\mathbf a_k,\boldsymbol{\omega}_k\}$
\Ensure Motion increment $\mathbf u_m=(s_m,\psi_m)$

\State Initialize quaternion $\mathbf q_0$, gyro bias $\boldsymbol{\eta}_0$, and $\mathcal T\gets\emptyset$
\For{each sample $k$}
    \State \textbf{StanceDetect:} $r_k=\mathcal S(\mathbf a_{1:k})$
    \State \textbf{BiasUpdate:} $\boldsymbol{\eta}_k=\mathcal G(\boldsymbol{\omega}_{1:k},\mathbf a_{1:k})$
    \State \textbf{Heading:} 
    $\mathbf q_k=\mathcal H(\mathbf q_{k-1},\mathbf a_k,\boldsymbol{\omega}_k-\boldsymbol{\eta}_k)$, 
    $\psi_k=\Psi(\mathbf q_k)$
    \If{$r_k=1$}
        \State Append $k$ to $\mathcal T$
        \If{two stance onsets $\tau_{m-1},\tau_m\in\mathcal T$ available}
            \State \textbf{Output:} 
            $s_m=\mathcal D(\mathbf a_{\tau_{m-1}:\tau_m})$, 
            $\mathbf u_m=(s_m,\psi_{\tau_m})$
        \EndIf
    \EndIf
\EndFor
\end{algorithmic}
\end{minipage}%
}
\end{algorithm}

\subsection{Positioning Server}
The positioning server serves as the system’s central processing engine. To support multiple users, it uses a distributed worker architecture, where each processing worker handles one wearable tag and maintains its own parameters and state variables for consistent trajectory estimation. When a packet arrives, the server retrieves the corresponding building context and indoor layouts from the map database, then executes the Map-Aware Particle Filter (MAPF). By offloading intensive fusion to the server, the system can jointly integrate GNSS, PDR-based motion updates, and pressure for seamless outdoor--indoor tracking and floor-change detection without manual mode switching. The server also performs map matching against walls and entrance constraints, and executes floor detection while maintaining real-time responsiveness. Estimated positions are stored for safety and performance analytics.

MAPF replaces the fixed predict–update–resample cycle of conventional particle filters with an event-driven design that updates whenever motion, GNSS, or pressure data arrive, matching the system’s asynchronous sensing flow. Algorithm~\ref{alg:map_pf} summarizes the procedure. Here, $\mathcal{C}(\cdot)$ enforces wall-collision and entrance-based outdoor--indoor transition constraints; $\alpha_k=C(\sigma_h)S_v$ weights GNSS updates using horizontal accuracy and velocity consistency; $\mathcal{B}(\cdot)$ detects vertical transitions using DBSCAN and estimates the current floor $l_k$ and transition context $q_k$ using an SVM-based classifier, following~\cite{sung2022floor}. Then, $\mathcal{T}(\cdot)$ updates the particles and loads the occupancy map of the corresponding floor.

\begin{algorithm}[t!]
\caption{Event-Driven Occupancy Map-Aware PF}
\label{alg:map_pf}
\resizebox{75mm}{!}{%
\begin{minipage}{\linewidth}
\begin{algorithmic}[1]
\Require Particles $\mathcal{P}_{k-1}=\{(\mathbf{x}_{k-1}^i,w_{k-1}^i)\}_{i=1}^{N}$, map $\mathcal{M}$
\Require Motion $\mathbf{u}_k$, GNSS $\mathbf{z}_k$, barometric pressure $\mathbf{b}_k$
\Ensure State estimate $\hat{\mathbf{x}}_k$

\For{each incoming event at time $k$}
    \If{motion $\mathbf{u}_k$ arrives}
        \For{$i=1$ to $N$}
            \State \textbf{Propagate:} 
            $\mathbf{x}_k^i \leftarrow f(\mathbf{x}_{k-1}^i,\mathbf{u}_k)+\boldsymbol{\epsilon}_k$
            \State \textbf{MapCheck:} 
            $w_k^i \leftarrow w_{k-1}^i\mathcal{C}(\mathbf{x}_{k-1}^i,\mathbf{x}_k^i;\mathcal{M})$
        \EndFor

    \ElsIf{GNSS $\mathbf{z}_k$ arrives}
        \State \textbf{Likelihood:} 
        $\tilde{\mathbf{w}}_k \leftarrow p(\mathbf{z}_k|\mathcal{P}_k)$; 
        $\alpha_k \leftarrow C(\sigma_h)S_v$
        \State \textbf{GNSSUpdate:} 
        $\mathbf{w}_k \leftarrow (1-\alpha_k)\mathbf{w}_{k-1}+\alpha_k\tilde{\mathbf{w}}_k$

    \ElsIf{pressure $\mathbf{b}_k$ arrives}
        \State \textbf{BaroUpdate:} 
        $(l_k,q_k) \leftarrow \mathcal{B}(\mathbf{b}_{1:k})$; 
        $\mathcal{M} \leftarrow \mathcal{M}(l_k)$
        \State \textbf{ContextUpdate:} 
        $\mathcal{P}_k \leftarrow \mathcal{T}(\mathcal{P}_k,l_k,q_k;\mathcal{M})$
    \EndIf

    \State \textbf{Normalize:} 
    $\mathbf{w}_k \leftarrow \mathbf{w}_k/\sum_i w_k^i$; 
    $\hat{\mathbf{x}}_k \leftarrow \sum_i w_k^i\mathbf{x}_k^i$

    \If{$N_{\mathrm{effective}} < N_{\mathrm{threshold}}$}
        \State \textbf{Resample:} 
        $\mathcal{P}_k \leftarrow \mathrm{Resample}(\mathcal{P}_k)$
    \EndIf
\EndFor
\end{algorithmic}
\end{minipage}%
}
\end{algorithm}

\subsection{Database}

The proposed LBSM system uses two connected MongoDB repositories: a general database and an indoor-map database. Both interface with the positioning and dashboard servers, supporting seamless data exchange across storage, localization, and visualization. The general database stores dynamic operational data, including user profiles, device records, sensor measurements, and positioning and context outputs used for worker tracking and daily operation. The indoor-map database stores spatial data for map-aware fusion and visualization, such as building metadata, floor plans, occupancy maps, reachable areas, landmarks, access types, directions, and spatial relationships. It also connects to the map tool, allowing verified map data to be uploaded and used directly by the system.


\subsection{Dashboard Server}

\subsubsection{Real-time Monitoring}

\begin{figure}[t!]
    \centering
      \includegraphics[width = 85 mm]{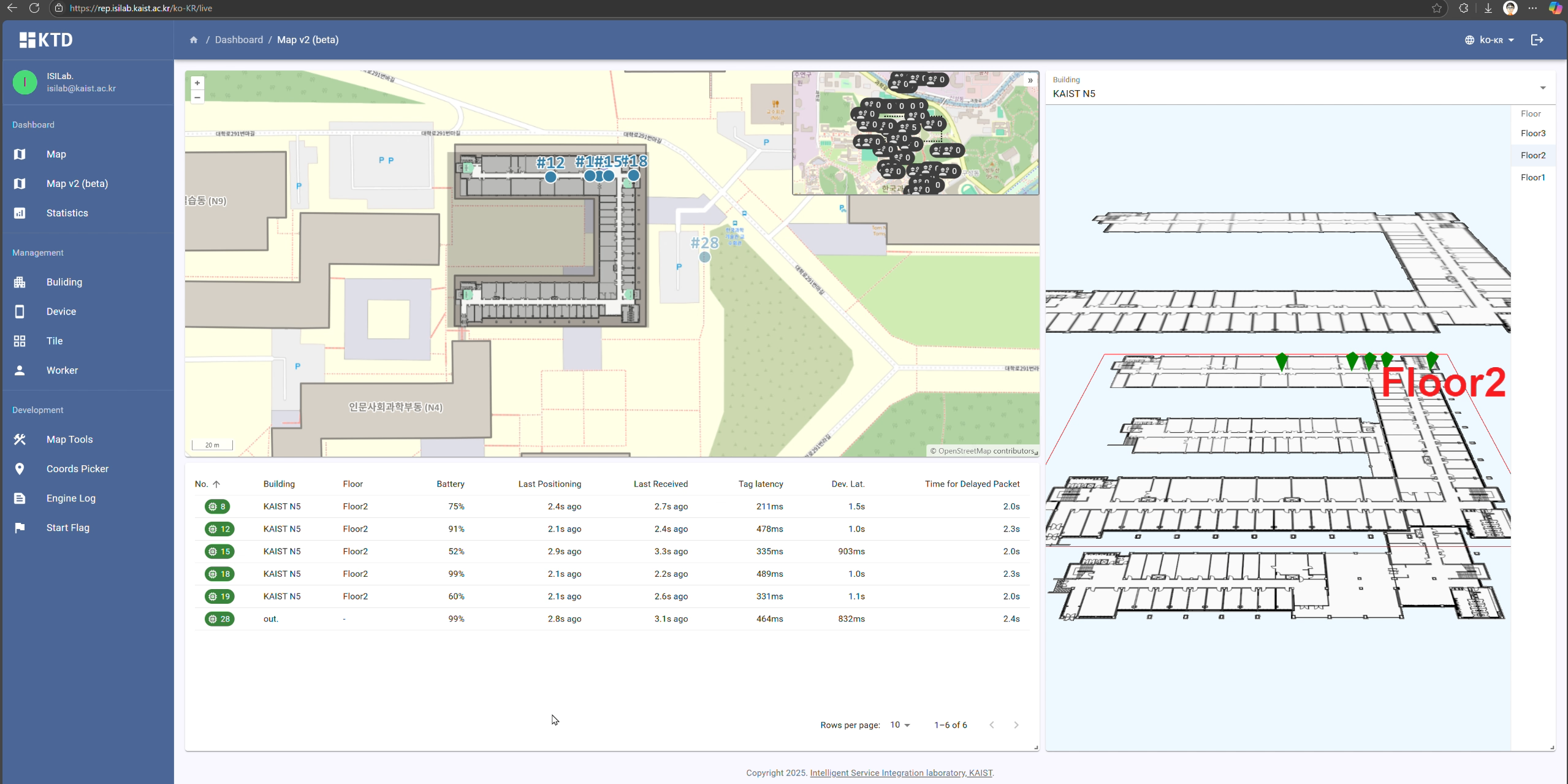}
      \caption{Real-time safety management dashboard via the web platform.}
      \label{fig:monitoring_system}
\end{figure}

Site managers need a real-time view of worker locations and system status. The dashboard provides a unified web interface for device registration, worker assignment, and live monitoring across indoor and outdoor spaces. To support intuitive supervision, it visualizes the worksite in both 2D and 3D, helping managers quickly understand worker distribution, movement, and spatial context. It also reports key operational states, including device connectivity, network latency, battery level, current building and floor, allowing inactive or malfunctioning wearable tags to be detected promptly. In addition, the dashboard summarizes indoor and outdoor tag counts by building for efficient site-wide monitoring. The system is implemented as a full-stack web app using Next.js and retrieves visualization data from the general database. As shown in Fig.~\ref{fig:monitoring_system}, once a tag is active, it appears as a green icon in the status panel, with its position displayed across the site, floor, and multi-floor map views.

\subsubsection{Location-based Analytics}
The dashboard server also provides a location-based analytics module for safety and performance reporting. The tool transforms recorded location data into actionable insights over a selected period, including active worker counts, travel distance, movement and idle duration, inter-floor transitions, geofence activity, and hazardous-zone entries. These metrics help managers assess worker mobility, site activity, and potential safety risks. Accumulated traces can also reveal behavioral patterns, such as collaboration through overlapping paths or shared work areas. As monitoring needs evolve, the module can incorporate additional indicators and automatically consolidate daily reports into weekly and monthly summaries for long-term trend analysis.

\subsubsection{Geofencing }

Geofencing is a key LBSM feature that supports accident prevention and improves situational awareness in workplaces. Administrators can create, update, or remove hazardous and restricted zones, either by drawing boundaries manually or selecting automatically generated rectangles or hexagons. Each zone can include predefined actions and warning messages, enabling immediate and consistent responses when a worker enters or approaches a risky area. Once a tag is detected inside a configured zone, alerts are delivered through the tag or mobile application, helping workers recognize hazards and enabling managers to respond quickly. Beyond safety alerts, it can also provide operational guidance and contextual notifications, extending geofencing to broader location-aware worksite services.

\subsection{Map Tool}
The map tool converts floor-plan sketches into operational indoor maps for the system. It provides a semi-automatic workflow in which administrators import floor plans, align them with real-world coordinates, and annotate accessible areas, landmarks, entrances, stairs, elevators, and escalators. The tool generates a binary occupancy map used by the particle filter for particle propagation, collision checking, and floor-transition handling, as well as a visual map representation for visualization and monitoring. It also stores metadata on floor structure, reachable areas, landmark groups, access types, directions, and spatial relationships, transforming a static floor image into a navigable spatial model of the indoor environment. The generated map data are stored in the indoor-map database and shared across the platform. Technical details of the map tool are presented in~\cite{son2025framework}.


\section{Evaluation}
\subsection{Experimental Setup}

In the experiments, ten participants wore the devices on the front of their chests, which connected to the server via LTE Cat-M. The IMU, barometer, and GNSS were sampled at 100~Hz, 10~Hz, and 1~Hz, respectively, and the global fusion module used $N=1500$ particles. Each participant walked along preplanned $\sim$1400 m trajectories covering both outdoor and indoor areas. Outdoor ground truth was obtained from RTK GNSS logs. For indoor segments, we generated map-annotated reference trajectories by geo-registering official floor plans into a global coordinate frame. Each floor plan was aligned with OpenStreetMap building outlines using RTK-surveyed non-collinear corners.

\subsection{Positioning Evaluation}

Table~\ref{tab:positionig_acc} compares our system with representative positioning approaches. While prior systems achieve strong accuracy in controlled indoor settings, many depend on smartphones, local processing, post-processing, or BLE/UWB infrastructure. In contrast, our implemented system provides competitive accuracy over a $\sim$1400~m outdoor--indoor testbed using a resource-constrained wearable and a device--cloud split architecture. It achieves 1.32~m mean error and 1.80~m RMSE indoors, comparable to infrastructure-based indoor systems, while maintaining robust outdoor accuracy with 3.30~m mean error and 3.64~m RMSE. The performance of MAPF with other configurations is presented in Table~\ref{tab:MAPF_ablation}. These results show that the proposed architecture supports practical, continuous positioning across heterogeneous environments under wearable resource constraints. A demo video is available at~\cite{youtube_4kyJpvb1ukc}.

\begin{table*}[t!]
\renewcommand{\arraystretch}{1.3}
\centering
\caption{Comparison of positioning accuracy between systems}
\label{tab:positionig_acc}
\resizebox{\textwidth}{!}{%
\begin{tabular}{|c|c|c|c|c|ccc|ccc|}
\hline
\multirow{2}{*}{\textbf{Approaches}} &
  \multirow{2}{*}{\textbf{Input Data}} &
  \multirow{2}{*}{\textbf{Testbed}} &
  \multirow{2}{*}{\textbf{\begin{tabular}[c]{@{}c@{}}Sensing Platform\end{tabular}}} &
  \multirow{2}{*}{\textbf{\begin{tabular}[c]{@{}c@{}}System Structure\end{tabular}}} &
  \multicolumn{3}{c|}{\textbf{Indoor (m)}} &
  \multicolumn{3}{c|}{\textbf{Outdoor (m)}} \\ \cline{6-11} 
 &
   &
   &
   &
   &
  \multicolumn{1}{l|}{\textbf{Mean}} &
  \multicolumn{1}{l|}{\textbf{Std.}} &
  \multicolumn{1}{l|}{\textbf{RMSE}} &
  \multicolumn{1}{l|}{\textbf{Mean}} &
  \multicolumn{1}{l|}{\textbf{Std.}} &
  \multicolumn{1}{l|}{\textbf{RMSE}} \\ \hline
Han et al. (2014) \cite{han2014kailos} &
  Wi-Fi/IMU &
  Indoor &
  Smartphone &
  Cloud Offloading &
  \multicolumn{1}{c|}{1.98} &
  \multicolumn{1}{c|}{1.48} &
  - &
  \multicolumn{1}{c|}{-} &
  \multicolumn{1}{c|}{-} &
  - \\ \hline
Chen et al. (2021) \cite{chen2021data}&
  BLE/IMU &
  Indoor (52.5x52.5m) &
  Smartphone &
  Post Processing &
  \multicolumn{1}{c|}{1.01} &
  \multicolumn{1}{c|}{-} &
  - &
  \multicolumn{1}{c|}{-} &
  \multicolumn{1}{c|}{-} &
  - \\ \hline
Hu et al. (2022) \cite{hu2022experience}&
  BLE/IMU &
  Indoor Shopping Mall &
  Smartphone &
  Cloud Offloading &
  \multicolumn{1}{c|}{3.10} &
  \multicolumn{1}{c|}{-} &
  - &
  \multicolumn{1}{c|}{-} &
  \multicolumn{1}{c|}{-} &
  - \\ \hline
Mayer et al. (2023)\cite{mayer2023self}&
  UWB &
  Indoor (200 m2 room) &
  Custom Hardware &
  Local Processing &
  \multicolumn{1}{c|}{1.20} &
  \multicolumn{1}{c|}{-} &
  - &
  \multicolumn{1}{c|}{-} &
  \multicolumn{1}{c|}{-} &
  - \\ \hline
Wang et al. (2023) \cite{wang2023seamless}&
  GNSS PPP-3DMA/IMU &
  Indoor/Outdoor (\textless{}600 m) &
  Custom Hardware/Smartphone &
  Local Processing &
  \multicolumn{1}{c|}{-} &
  \multicolumn{1}{c|}{-} &
  - &
  \multicolumn{1}{c|}{-} &
  \multicolumn{1}{c|}{-} &
  0.72 \\ \hline
Zhang et al. (2025) \cite{zhang2025seamless}&
  GNSS/IMU/UWB/Map &
  Outdoor/Indoor &
  Smartphone &
  Local Processing &
  \multicolumn{1}{c|}{3.10} &
  \multicolumn{1}{c|}{1.87} &
  3.91 &
  \multicolumn{1}{c|}{4.76} &
  \multicolumn{1}{c|}{5.50} &
  10.77 \\ \hline
Ours &
  GNSS/IMU/Pressure/Map &
  Outdoor/Indoor ($\sim$1400 m) &
  Custom Wearable &
  Device/Cloud Split &
  \multicolumn{1}{c|}{1.32} &
  \multicolumn{1}{c|}{1.23} &
  1.80 &
  \multicolumn{1}{c|}{3.30} &
  \multicolumn{1}{c|}{1.53} &
  3.64 \\ \hline
\end{tabular}%
}

\vspace{2pt}
\begin{minipage}{0.98\textwidth}
\fontsize{7.0}{9}\selectfont
\textit{Note: Zhang et al.’s results are from the iQOO smartphone.; Wang et al. report a 1.17 m RMSE over the full indoor–outdoor trajectory.}
\end{minipage}
\end{table*}

\begin{table}[t!]
    \centering
    \caption{Ablation study on the positioning accuracy of MAPF}
    \label{tab:MAPF_ablation}
    \resizebox{90mm}{!}{
    \begin{tabular}{
        c l
        S[table-format=2.2, detect-weight]
        S[table-format=2.2, detect-weight]
    }
        \toprule
        \textbf{Environment} & \textbf{Method} & {\textbf{Mean (m)}} & {\textbf{RMSE (m)}} \\
        \midrule
        \multirow{4}{*}{Outdoor}      & GNSS                        & 4.71           & 6.87           \\
                                      & MAPF (No indoor map)        & 3.38           & 3.85           \\
                                      & MAPF (No velocity consistency weighting)         & 3.35           & 3.68           \\
                                      & MAPF                        & \bfseries 3.30 & \bfseries 3.64 \\
        \midrule
        \multirow{4}{*}{Indoor}       & GNSS                        & 22.22          & 22.96          \\
                                      & MAPF (No indoor map)        &  6.98          &  7.94          \\
                                      & MAPF (No velocity consistency weighting)         &  1.39          &  2.04          \\
                                      & MAPF                        & \bfseries 1.32 & \bfseries 1.80 \\
        \bottomrule
    \end{tabular}
    }

\end{table}

\subsection{Positioning Latency Evaluation}

\begin{figure}[t!]
  \centering
  \includegraphics[width = 80mm]{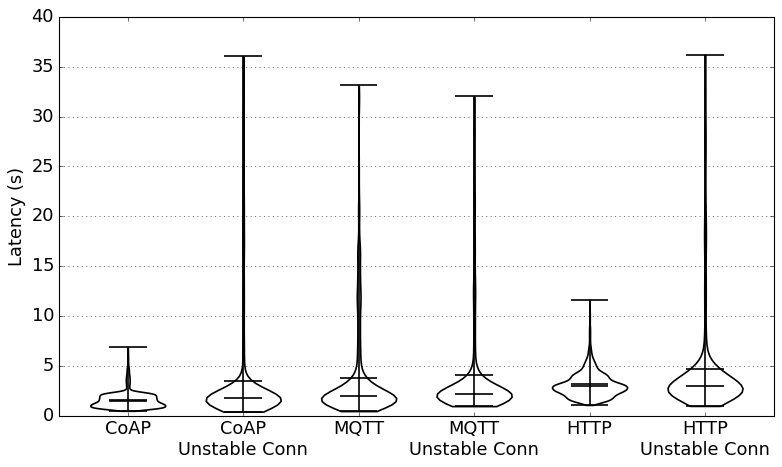}
  \caption{Positioning latency under normal and unstable connectivity.}
  \label{fig:positioning_laterncy}
\end{figure}

Fig.~\ref{fig:positioning_laterncy} compares positioning latency for three IoT protocols under normal and degraded LTE Cat-M connectivity. Intermittent outages were emulated by disabling the modem every 1--3 minutes for 10--30 seconds. Under normal conditions, CoAP achieves the lowest latency, with a mean of 1.62~s, a median of 1.48~s, and a 95th percentile of 3.23~s. HTTP remains stable at a mean of 3.16~s and a 95th percentile of 5.50~s, while MQTT exhibits greater variability, reaching 3.80~s and 15.35~s, respectively. Under degraded connectivity, median latency increases only slightly, but tail latency rises sharply. The 95th percentiles of CoAP and HTTP increase to 18.43~s and 18.84~s, whereas MQTT shows the smallest degradation and the lowest value of 18.09~s. Overall, the system maintains low typical latency over IoT-grade links, although protocol choice significantly affects robustness to intermittent connectivity.

\subsection{Geofencing-Based Alert Evaluation}
To verify the safety-management capability of the system, we evaluated geofence-based access detection using 10 predefined restricted zones. Ten participants repeatedly entered the predefined zones, producing 100 zone-access events in total. The system detected all zone-entry events, corresponding to a 100\% entry-detection rate with no missed entries in this test. The platform-side alert-generation latency, measured from zone entry to alert generation on the safety-management platform, was 1.76 s. These results indicate that the proposed system allows timely and reliable location-based access alerts for practical safety monitoring.

\subsection{Power Consumption Evaluation}

We measured the wearable tag’s power consumption using Nordic’s Power Profiler Kit II. As shown in Table~\ref{tab:power_consumption}, HTTP draws the lowest average current of 155.6 mA and peak currents of 335.3 mA, while CoAP shows a moderate average current draw of 160.7 mA and peak current of 433.7 mA. In contrast, MQTT records the highest average current of 167.5 mA and peak current of 640.0 mA, reflecting more pronounced short-term communication bursts. With a 1800 mAh battery, the device provides approximately 10–12 h of operation, depending on network and environmental conditions.

\begin{table}[t!]
\renewcommand{\arraystretch}{1.3}
\centering
\caption{Power consumption of the wearable tag}
\label{tab:power_consumption}
\resizebox{80mm}{!}{%
\begin{tabular}{@{}ccc@{}}
\toprule
\textbf{Used Protocol} & \textbf{Average Current (mA) @4.2V} & \textbf{Peak Current (mA) @4.2V}  \\ \midrule
CoAP                    & 160.7                             & 433.7                         \\
HTTP                    & 155.6                             & 335.3                         \\
MQTT                    & 167.5                             & 640.0                        \\ \bottomrule
\end{tabular}%
}
\end{table}

\section{Conclusion and Future Work}

This work implemented a practical seamless positioning system for LBSM using a resource-constrained wearable tag and a device--cloud split architecture. The tag integrates GNSS, IMU, barometric sensing, LTE Cat-M, and lightweight local processing, while the server performs map-aware fusion for continuous outdoor--indoor tracking. Building on this positioning capability, the system further supports map management, real-time monitoring, geofencing-based access alerts, and safety/performance analytics. Experimental results show that the system offers a practical balance among accuracy, latency, and resource efficiency for scalable LBSM. Future work will enhance robustness in challenging deployments by exploring NPU-assisted data-driven PDR on the wearable tag, improving energy efficiency and thermal stability for long-term use, and expanding safety-management features. These efforts will move the platform toward continuous, scalable, and field-deployable safety management.

\bibliographystyle{IEEEtran}
\bibliography{references}

\end{document}